\documentclass[letterpaper, 10 pt, conference]{ieeeconf}  %

\IEEEoverridecommandlockouts                              %

\makeatletter
\let\NAT@parse\undefined  %
\makeatother

\usepackage[bottom]{footmisc}
\usepackage{dblfloatfix}
\usepackage{gensymb}
\usepackage{graphicx}
\usepackage{hyperref}
\usepackage{lipsum}
\usepackage{subcaption}
\usepackage{textcomp}
\usepackage{url}
\usepackage{xcolor}
\usepackage{multirow}
\usepackage{array}

\title{\LARGE \bf
    Towards an Extremely Robust Baby Robot With Rich Interaction Ability for Advanced Machine Learning Algorithms
}

\author{
    Mohannad Alhakami,\hspace{-0.025em}$^{*}$\hspace{0.05em}$^{1,2}$ Dylan R.~Ashley,\hspace{-0.025em}$^{*}$\hspace{0.05em}$^{3,4,5,6}$ Joel Dunham,\hspace{-0.025em}$^{*}$\hspace{0.05em}$^{7}$ Yanning Dai,$^{3}$\\
    Francesco Faccio,$^{3,4,5,6}$ Eric Feron,$^{1}$ and J\"{u}rgen Schmidhuber\hspace{0.2em}$^{3,4,5,6,8}$
    \thanks{$^{*}$Equal contribution. $^{1}$Robotics, Intelligent Systems, and Control Lab, King Abdullah University of Science and Technology (KAUST), Saudi Arabia. $^{2}$Saudi Basic Industries Corporation (SABIC), Saudi Arabia. $^{3}$Center of Excellence in Generative AI, King Abdullah University of Science and Technology (KAUST), Saudi Arabia. $^{4}$Dalle Molle Institute for Artificial Intelligence Research (IDSIA), Switzerland. $^{5}$Universit\`{a} della Svizzera italiana (USI), Switzerland. $^{6}$Scuola universitaria professionale della Svizzera italiana (SUPSI), Switzerland. $^{7}$optoXense, Inc., United States of America. $^{8}$NNAISENSE, Switzerland. Correspondence to \href{mailto:mohannad.alhakami@kaust.edu.sa}{\tt mohannad.alhakami@kaust.edu.sa}}
}

\begin{document}

\bstctlcite{IEEEexample:BSTcontrol}

\maketitle
\thispagestyle{empty}
\pagestyle{empty}

\setcounter{footnote}{8}.  %

\begin{abstract}

Advanced machine learning algorithms require platforms that are extremely robust and equipped with rich sensory feedback to handle extensive trial-and-error learning without relying on strong inductive biases. Traditional robotic designs, while well-suited for their specific use cases, are often fragile when used with these algorithms. To address this gap---and inspired by the vision of enabling curiosity-driven baby robots---we present a novel robotic limb designed from scratch. Our design has a hybrid soft-hard structure, high redundancy with rich non-contact sensors (exclusively cameras), and easily replaceable failure points. Proof-of-concept experiments using two contemporary reinforcement learning algorithms on a physical prototype demonstrate that our design is able to succeed in a simple target-finding task even under simulated sensor failures, all with minimal human oversight during extended learning periods. We believe this design represents a concrete step toward more tailored robotic designs for achieving general-purpose, generally intelligent robots.

\end{abstract}

\section{INTRODUCTION} \label{sec:introduction}

\begin{figure}[p]
    \centering
    \begin{subfigure}[t]{0.75\linewidth}
        \includegraphics[width=\linewidth]{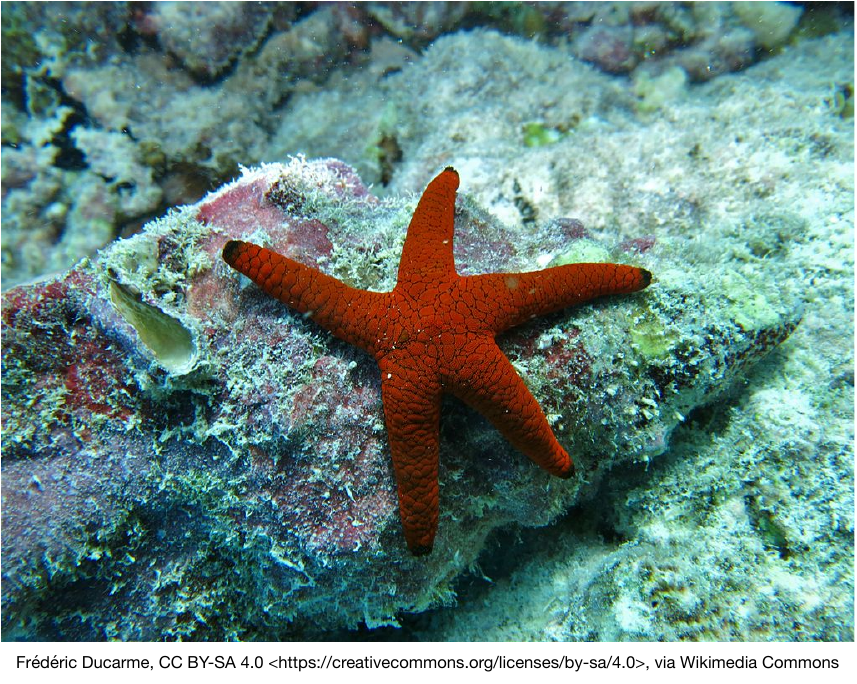}
        \caption{The principal inspiration for the design.\vspace{1em}}
        \label{fig:starfish}
    \end{subfigure}
    \begin{subfigure}[t]{0.75\linewidth}
        \includegraphics[width=\linewidth]{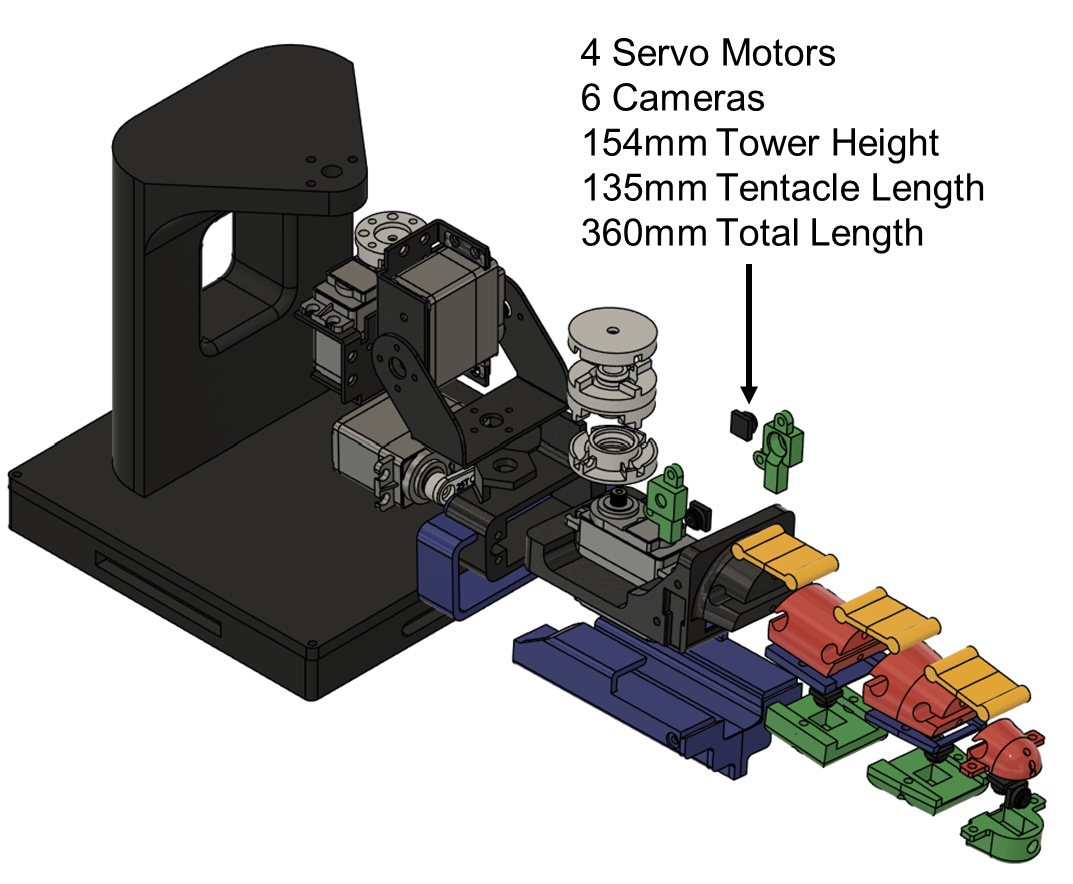}
        \caption{The final limb design.\vspace{1em}}
        \label{fig:iso_view}
    \end{subfigure}
    \begin{subfigure}[t]{0.75\linewidth}
        \includegraphics[width=\linewidth]{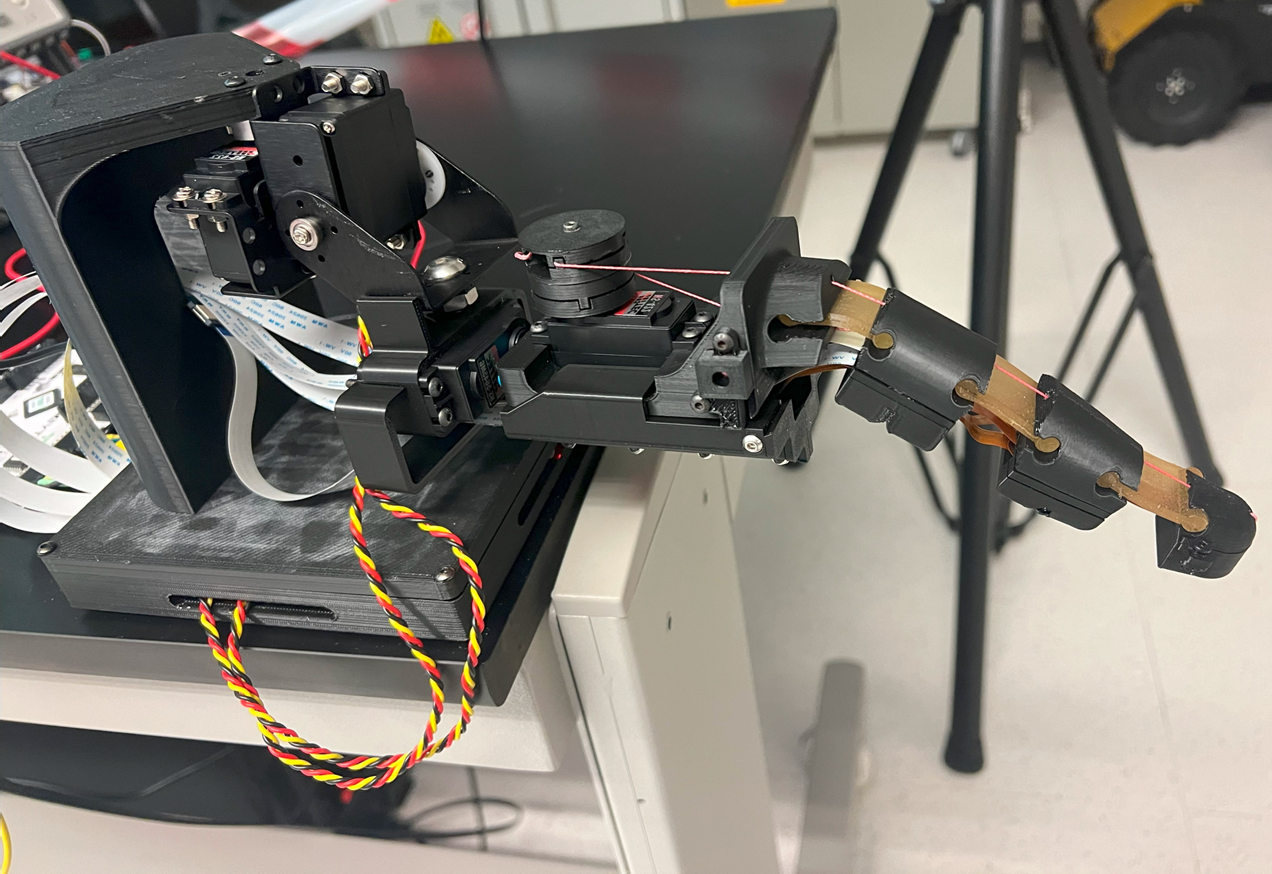}
        \caption{The constructed limb.}
        \label{fig:iso_photo}
    \end{subfigure}
    \caption{The design of the limb we construct is loosely inspired by the shape of a starfish in that we envision a multi-limbed headless creature with the bulk of the sensors being contained in the limbs themselves.}
    \label{fig:robot}
\end{figure}

Artificial intelligence (AI)---and in particular curiosity-driven baby robots---has long been touted as a way to achieve skilled general-purpose robots by enabling them to learn abstract knowledge about the world~\cite{schmidhuber1991curious,schmidhuber2006developmental}.
However, AI's potential in robotics has been severely hampered by the unsuitability of traditional robotic design to some of the more advanced machine learning (ML) algorithms \cite{korenkevych2019autoregressive,peters2007reinforcement}.
With the recent high-profile successes of AI in other areas (e.g., in generating text and images~\cite{openai2022introducing,ramesh2021zero,ramesh2022hierarchical,betker2023improving}), protein folding~\cite{jumper2021highly}, circuit design~\cite{roy2021prefixrl}, navigation of high-altitude balloons~\cite{bellemare2020autonomous}, controlling plasma in a tokamak reactor~\cite{degrave2022magnetic}, and achieving grandmaster level in Starcraft II~\cite{vinyals2019grandmaster}), there is reason to believe that the difficulties faced by robotics when attempting to leverage advanced ML algorithms may very well be holding the field back from a breakthrough.
To address this, we return to the fundamentals and undertake the cumbersome task of designing and building a robotic limb from scratch specifically for such algorithms.

Our design is focused on leveraging the properties most needed by advanced ML algorithms. These algorithms, particularly neural networks (NNs), are most effective when the platform on which they are deployed is \textbf{(1)}~extremely robust to handle extensive periods of trial-and-error learning and \textbf{(2)}~has rich sensory feedback and actuation potential to ensure that any desired task is achievable without the need for strong inductive biases.
In traditional robotics, the above are often considered undesirable traits, with most robot designs intended to carry out a very narrow set of smooth tasks using a minimal number of sensors and actuators.
Such designs are often easily damaged by small random movements as well as sub-optimal and destructive behaviour that characterize the learning process of advanced ML systems~\cite{korenkevych2019autoregressive,peters2007reinforcement}.

To allow for long periods of uninterrupted trail-and-error learning without damaging the robot, we focus on a design (shown in Figure~\ref{fig:iso_view}) that has (1)~a hybrid soft-hard design (with the soft area being the principal contact area), (2)~high redundancy in sensors, (3)~only non-contact sensors, and (4)~the most probable failure points located on parts designed to be easy to replace.
For sensors, we exclusively use cameras, as they provide particularly rich data streams and subsume most other kinds of common sensors (e.g., an autoregressive model on camera data can predict touch in most scenarios).
This design is loosely based on the shape of a starfish (see Figure~\ref{fig:starfish})---each limb is independent and, in contrast to an octopus, the contents of the central body are minimal.

We run a proof-of-concept experiment on a physical version of our design (shown in Figure~\ref{fig:iso_photo}) to demonstrate that it meets the aforementioned criteria.
This experiment uses both a state-of-the-art deep reinforcement learning algorithm (Proximal Policy Optimization) and a more classical algorithm in robotic control (Actor-Critic) with a simple target-finding task.
To demonstrate the utility of the redundant cameras, we experiment with randomly disabling one or more cameras in different episodes, simulating the effect of sensor failure.
In all cases, we show that both of these algorithms converge to optimal or near-optimal policies, all with minimal human supervision of the robot, despite extended runtimes.

Altogether, we believe that this robotic limb represents a concrete step forward toward a baby robot well-suited to applying advanced ML algorithms.
To help further that, we fully release the CAD files for all the robotic components, a complete implementation of the robot limb in the Gazebo simulator~\cite{koenig2004design}, and the source code for our experiments. The aforementioned can be found at \url{https://github.com/dylanashley/robot-limb-testai}

The main contributions of this paper are as follows:
\begin{itemize}
    \item Design a robotic limb to handle long-period reinforcement learning algorithms with minimal human oversight
    \item Design a robotic limb with redundant, rich sensors to enable machine learning even with degraded sensing
    \item Demonstrate two ML algorithms on the robotic limb with purposefully degraded sensing
\end{itemize}

The rest of the paper is organized as follows:
\begin{itemize}
    \item Section \ref{sec:background_and_related_work} details background and related work
    \item Section \ref{sec:mechanical_design} details the design methodology, materials selection, design, mechanical testing, and challenges of designing and building the limb
    \item Section \ref{sec:machine_learning_experiment} details ML experiments on this robotic limb
    \item Section \ref{conclusions_and_future_work} details conclusions and future improvements
\end{itemize}

\section{BACKGROUND AND RELATED WORK} \label{sec:background_and_related_work}

\begin{figure*}[b]
    \centering
    \includegraphics[width=.85\linewidth]{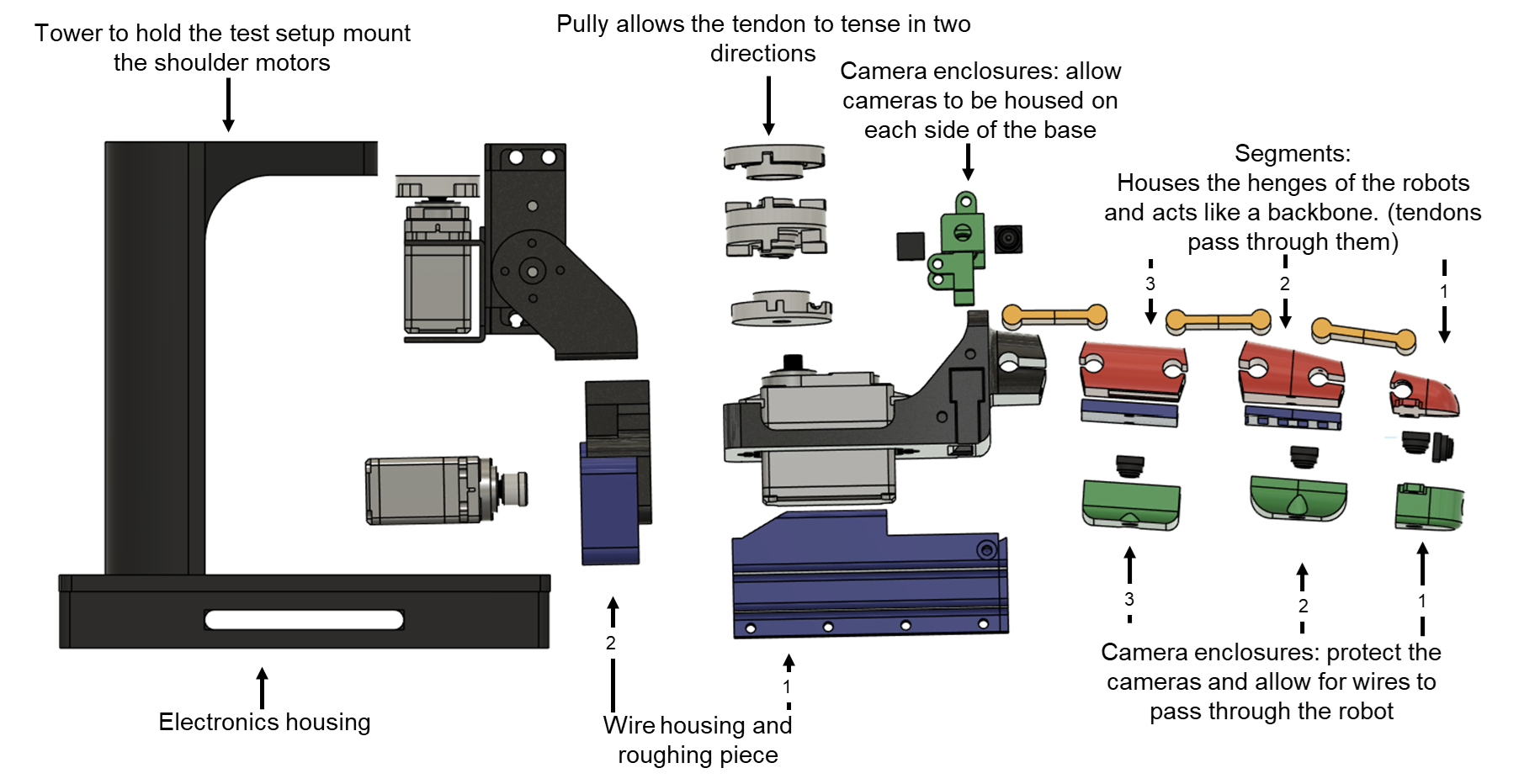}
    \caption{Exploded view of the Robot with colour-coding based on function---segments (red), brackets (black), camera enclosures (green), cable management (blue), and attachments(gray).}
    \label{fig:exploded_view}
\end{figure*}

Robotics has played a pivotal role in the development and popularization of reinforcement learning (RL), with many key achievements ranging from autonomous inverted helicopter flight~\cite{ng2006autonomous} to control of a complex robotic hand~\cite{andrychowicz2020learning}.
RL is thought to be most useful in learning a mapping from sensory and cognitive processes to movement, i.e., \textit{sensory-motor learning}~\cite{wolpert2001perspectives,peters2008natural,peters2007applying,degris2012model}.
Two particularly active areas of research for RL in robotics are in learning a predictive knowledge framework~\cite{littman2001predictive,modayil2014multi}, and for planning and control~\cite{ring2021representing}.
In the predictive knowledge framework in robotics, one of the most inspirational applications of RL is in prediction (e.g.,~\cite{pilarski2013adaptive}) and control (e.g.,~\cite{pilarski2013real}) of prosthetic limbs~\cite{pilarski2012dynamic,edwards2016application,edwards2016machine,vasan2017teaching}.
Today's AI-driven robotics environment is driven by relatively simple robot design requirements, e.g., a humanoid robot might be designed to demonstrate the AI's ability to meet a specific skill, such as playing basketball.
Many designs inspired by nature are likewise restricted to particular applications (e.g., a spider for mobility~\cite{rubio2019review}, an octopus for manipulation~\cite{calisti2011octopus}, or a humanoid for combining mobility and manipulation). Humanoid platforms such as iCub are designed to operate as research tools to test algorithms and mimic human behavior~\cite{beira2006design,tsagarakis2007icub}.

The design question to be answered---``What hardware attributes should a learning baby robot have in order to properly capture a large class of such systems?''---call upon the more recent research in incorporating flexibility in system design~\cite{saleh2009flexibility}.
Inspired by Turing's ``Child Machine''~\cite{turing1950computing}, traditional robots are constructed from rigid materials.
These materials, while aiding in repeatable and precise behavior, are susceptible to physical damage, especially on joints: bending or misaligned jeopardizes the robot's behavior.
Physical damage resistance can be improved by using soft robots even though there is a loss in dexterity and power~\cite{martinez2014soft}.
A scenario that is more favourable for us is a hybrid between rigid and soft components, where we replace the joints with compliant soft materials.
Yale soft hand is an example---a community-driven, low-cost robot using a hybrid manufacturing technique where 3D printed components are reinforced with soft, casted materials~\cite{ma2013modular}. A similar design was successfully implemented in the underwater humanoid robot, Ocean One, demonstrating its effectiveness in gripping and manipulation~\cite{stuart2017ocean}. Sensors are another critical component susceptible to damage in robotic systems, particularly during high-impact events. To mitigate this vulnerability. To address this vulnerability, increasing sensor redundancy can enhance fault tolerance and maintain functionality even in the event of sensor failures~\cite{albu2008soft}.

\section{ROBOTIC LIMB DESIGN} \label{sec:mechanical_design}

\paragraph{Methodology}
Similar to the robotic limb learning process, we designed the limb iteratively, learning from each set of experiments.
Low cost is a key requirement, facilitating rapid development and repair, since the limb was expected to fail as we increased robustness.
Commercial Off-The-Shelf (COTS) components were used whenever possible.
3D printed parts were used for the remaining mechanical parts.
Electronics were COTS and integrated through custom PCBs.
Due to the extensive use of 3D printing, material selection was a key step in the process.

\paragraph{Materials selection}
PLA---a widely used material with sufficient properties to enable initial testing---was used for rigid 3D printed parts as it is readily available, cheap, and fast to print.
The hinges of the soft portion of the arm are cast using flexible urethane rubber (smooth on-PMC\texttrademark-780 DRY) to prevent damage due to impact and compression commonly found in hinges constructed from traditional materials.
The two-part urethane material is cast using a reusable 3D printed mold, for fast iteration and cost-effective manufacturing.
A braided fishing line is used as a tendon to actuate the soft portion of the limb.
This fishing line is rated for 100 lb, exceeding expected requirements.
Kevlar thread was also tested; fishing line was selected due to availability.

\paragraph{Mechanical Design}
The limb is designed to behave in a natural starfish-like manner while housing and protecting the electronics in each segment.
The first iteration embedded the microcontrollers (MCUs) along the limb to keep the camera cables short.
However, this design quickly became self-limiting due to the number of required MCUs, resulting in a less flexible, more rigid, and far thicker limb.
Further, electronics embedded in the limb are difficult to replace, reducing the ability to rapidly develop a robust system.
Longer ribbon cables and motor wiring enabled all MCUs to be embedded in the base of the limb, significantly increasing flexibility and maintainability.

The critical components for flexiblity are the hinges, which are prone to damage.
We placed flexible hinges and tendons to enable flexing in two directions, and we designed segments and hinges such that the number of segments can be modified between robots without major redesigns.
Figure~\ref{fig:exploded_view} shows an exploded view of the robotic limb components.

Segments can be divided into three major types: the tip contains an enclosure housing two cameras; the connecting segment features one camera and a shape to ease the tendon route to the tip; the extension segment houses a camera and extends the limb length.
Those segments connect to the base, where two cameras are housed alongside the servo motor responsible for movement via tendon actuation.
The foregoing tendons are actuated using a custom pulley for bidirectional control.
The pulley clamps around each side of the tendon to avoid slippage.

This robotic limb provides an intermediate ground between a fully soft robot and a servo-controlled rigid joint limb.
As such, servos are used to act as a wrist and shoulder to provide more degrees of freedom and consistent response, enabling the soft part of the limb to be more fully utilized.
The combined motion of the actuators allows the limb to point the cameras and curl the soft portion in a semi-sphere, perform a grab motion in a $180\degree$ rotation from the main axis, move left/right in a $90\degree$ arc around the tower's vertical axis, and move in an up/down arc of approximately $90\degree$.

\begin{figure}[t]
    \centering
    \vspace{1em}
    \includegraphics[width=.95\linewidth]{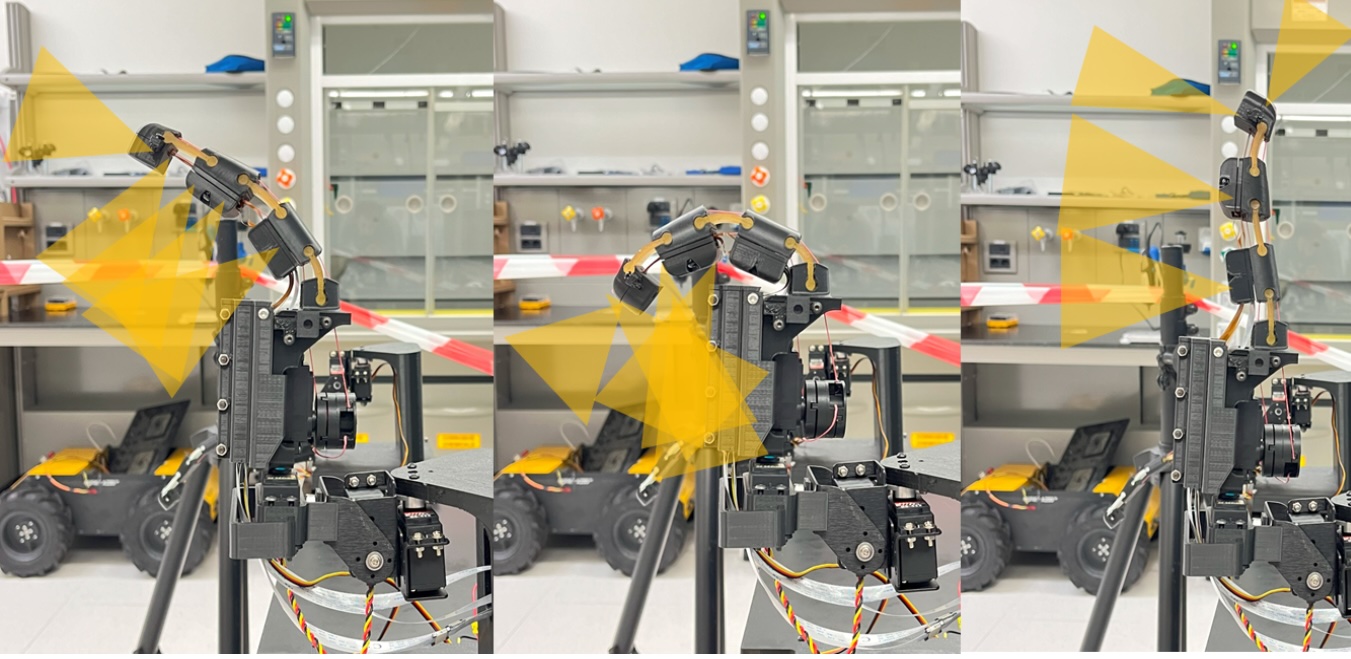}
    \caption{The effect on the field of view for the four front-facing cameras when the limb is extended and retracted.}
    \label{fig:camera_fields}
\end{figure}

\paragraph{Cameras and Sensors}
Cameras were chosen for all external-facing sensors, since they can act as a vision or contact sensor through the use of algorithms.
Placement of the cameras in the tip of the limb and sides of the base enables pointing and side tracking for vision-type tasks.
Cameras facing towards where the limb will touch objects (inside the curl of the limb and at the tip of the limb) enables contact sensing tasks.
Shells are shown in green in Figure~\ref{fig:exploded_view}.
Figure~\ref{fig:camera_fields} visualizes Field of Views (FOVs) in a different flexing of the soft portion of the limb.

\paragraph{Electronics Design}

\footnotetext[11]{See \url{https://digilent.com/shop/arty-a7-100t-artix-7-fpga-development-board/}}

\begin{figure}[b]
    \centering
    \begin{minipage}{0.525\linewidth}
        \includegraphics[width=\linewidth]{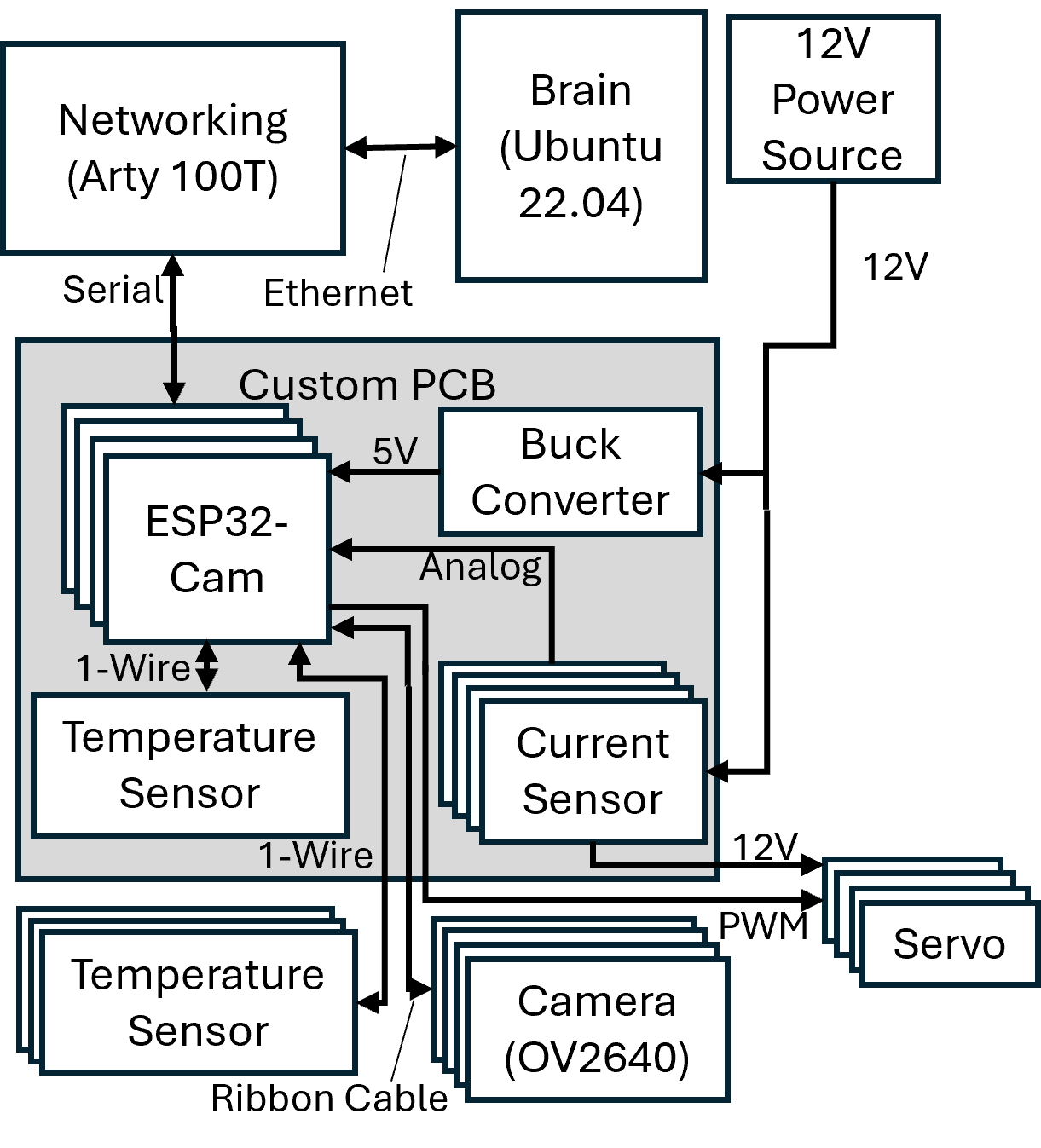}
        \caption[Wiring Diagram including the Arty A7 development board for networking.]{Wiring Diagram including the Arty A7 development board\footnotemark{} for networking.}
        \label{fig:pseudo_wiring}
    \end{minipage}
    \hfill
    \begin{minipage}{0.425\linewidth}
        \vspace{2.8em}
        \includegraphics[width=\linewidth]{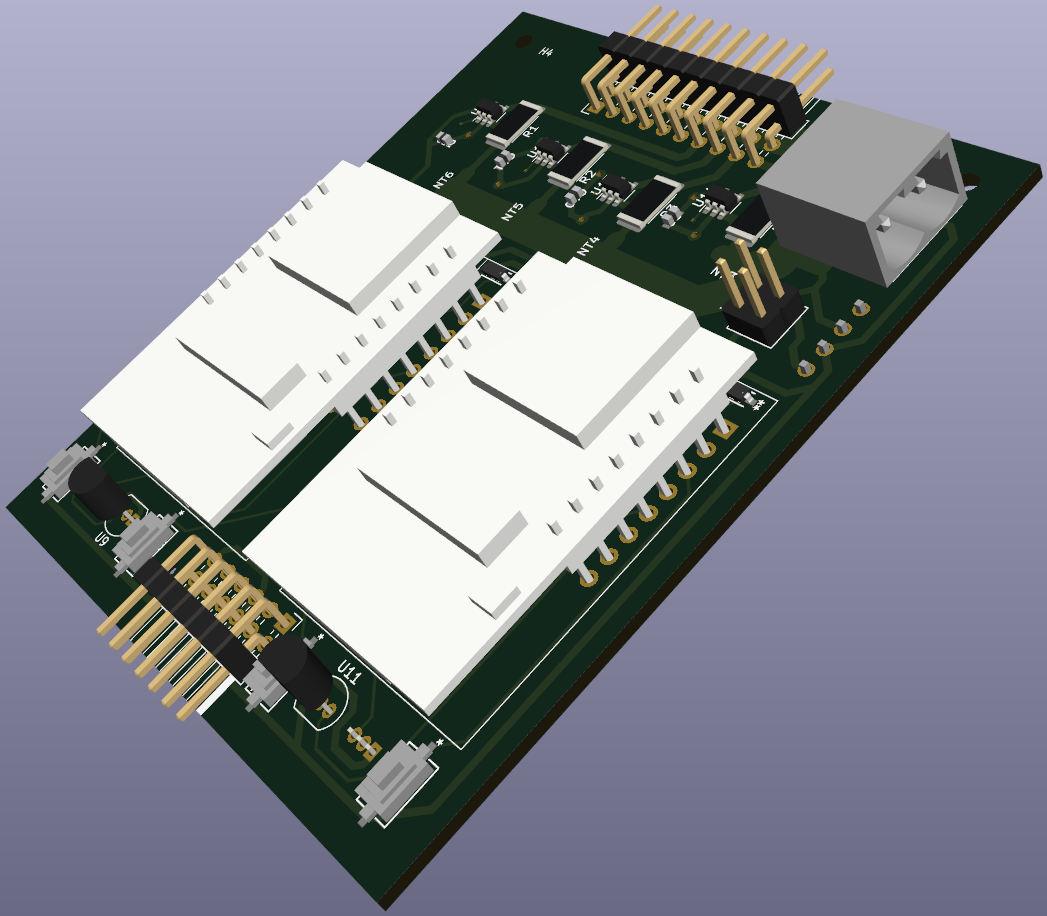}
        \caption{The custom PCB for combining ESP32-Cam(s), sensors, and power/networking wiring management.}
        \label{fig:pcb}
    \end{minipage}
\end{figure}

Electronics design for this robotic limb is challenging, in part due to the high number of camera sensors desired.
These sensors must transmit all data back to the brain for processing instead of edge processing on MCUs, resulting in high-bandwidth networking requirements.
We selected ESP32-Cam(s)\footnote{See \url{https://docs.ai-thinker.com/en/esp32-cam}}---cheap microprocessors that combine with OV2640\footnote{See \url{https://www.uctronics.com/download/cam\_module/OV2640DS.pdf}} vision sensors and provide WiFi and Bluetooth capabilities---as we can obtain them in high quantities for low costs.

The wireless networking of the ESP32s consumes significant resources, generates significant heat, and results in complicated IP management.
To avoid these issues, we chose to use physical wires, which also reduces heat by avoiding running the WiFi capabilities.
We integrated SatCat5~\cite{utter2020satcat5}---an open-source mixed-mode FPGA-based networking solution that was originally developed for use on cube satellites.
SatCat5 enables the networking to connect directly through serial, which is exposed on the ESP32-Cam(s).

We obtain up to 13 Frames per Second for 100x100 images, with the limitation being the ESP32-Cam processing, not the networking, enabling significant expansion of the number of cameras for later iterations.
Figure~\ref{fig:pseudo_wiring} shows the wiring diagram for a single limb, and Figure~\ref{fig:pcb} shows the custom PCB used to combine the ESP32-Cams, current sensors, and thermal sensors.

The current and thermal sensors enable the robot to "feel pain" and provide a surrogate for tactile/force feedback.
Embedded algorithms on the ESP32s provide safety mechanisms to avoid over-stressing the limb. With AI/ML algorithm improvement, the embedded algorithm limits will be gradually reduced, providing more flexibility to the AI/ML algorithms.

\paragraph{Wiring and Tendon Challenges}
Binding of wires is a common issue for robotic limbs.
Typically, this issue is solved via limiting motion to enable only specifically required tasks.
For an machine learning-focused robotic limb without \textit{a priori} known tasks, it is not possible to limit the motion this same way.
Therefore, we designed a two-layer passive tensioning attachment, wire housing, and a routing attachment to enable the wires to reach the base without hindering motion while protecting the cables.
This design was based on lessons learnt during the prototyping process, as loose ribbon cables were likely to get pinched and fail.
Key failure points include the camera extension cable connector, which will be adjusted in subsequent versions to avoid motion of that connector.

A second difficulty is the tendon connection to the servo; the tendon must avoid slipping while enabling actuation in both directions.
We created a pulley where the tendon is routed through teeth-like clamps to secure the tendon while minimizing risk of derailment.
While this method worked in a lab environment, the design ultimately provided information for future revisions to improve robustness in more free-form environments.

\paragraph{Mechanical Testing}
We performed mechanical testing focused on range of motion and ability of the robotic limb to perform ML experiments without harm.
We commanded each servo to full range, resulting in complete range testing and faster motion with harder stops than from ML commands.
Only the wrist axis required limiting to 150deg out of the full 180deg of servo-capable motion, due to cable binding at the extreme limits.
These limits ensured that long-period ML experiments without direct supervision could be performed without the robotic limb harming itself.

\begin{figure}[h]
    \centering
    \vspace{0.66em}
    \includegraphics[width=.9\linewidth]{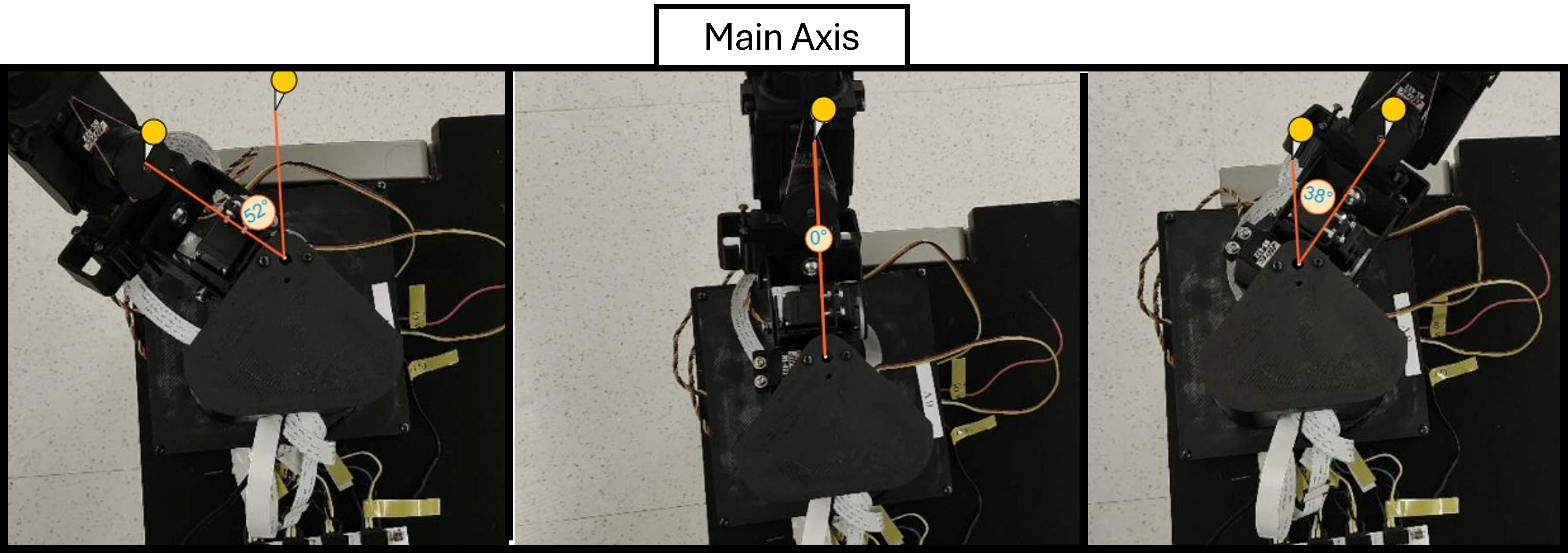}
    \includegraphics[width=.9\linewidth]{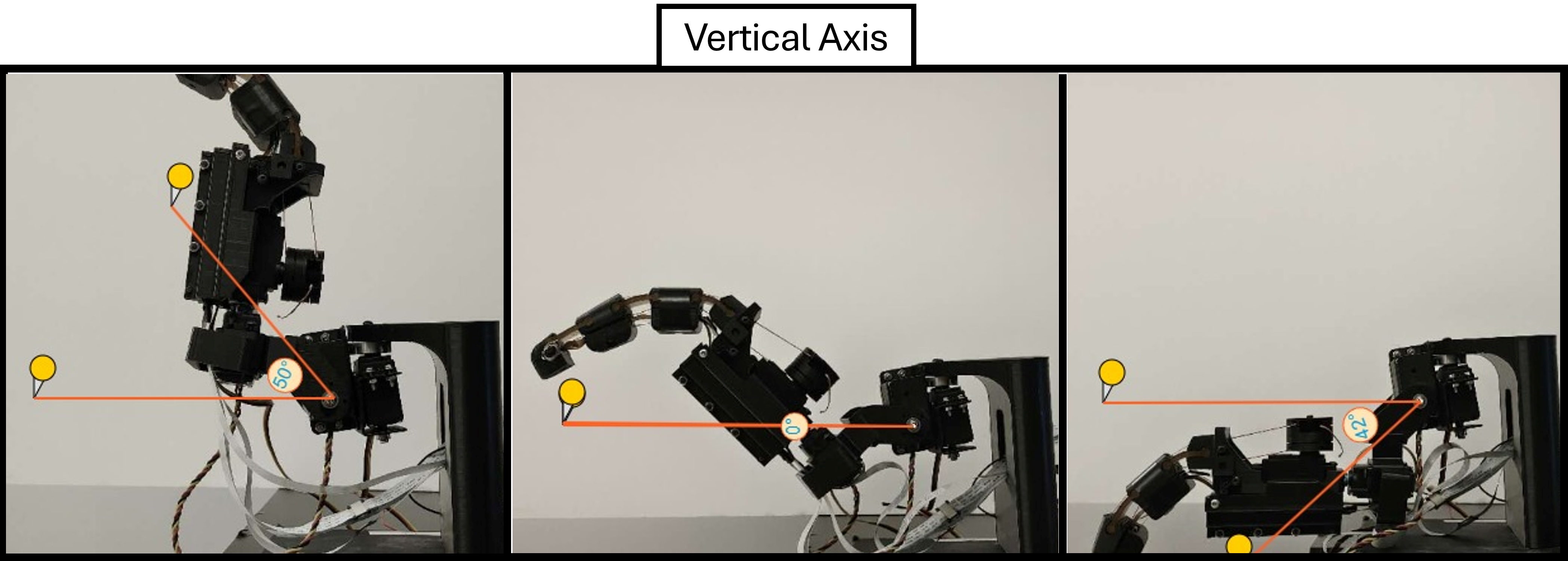}
    \includegraphics[width=.9\linewidth]{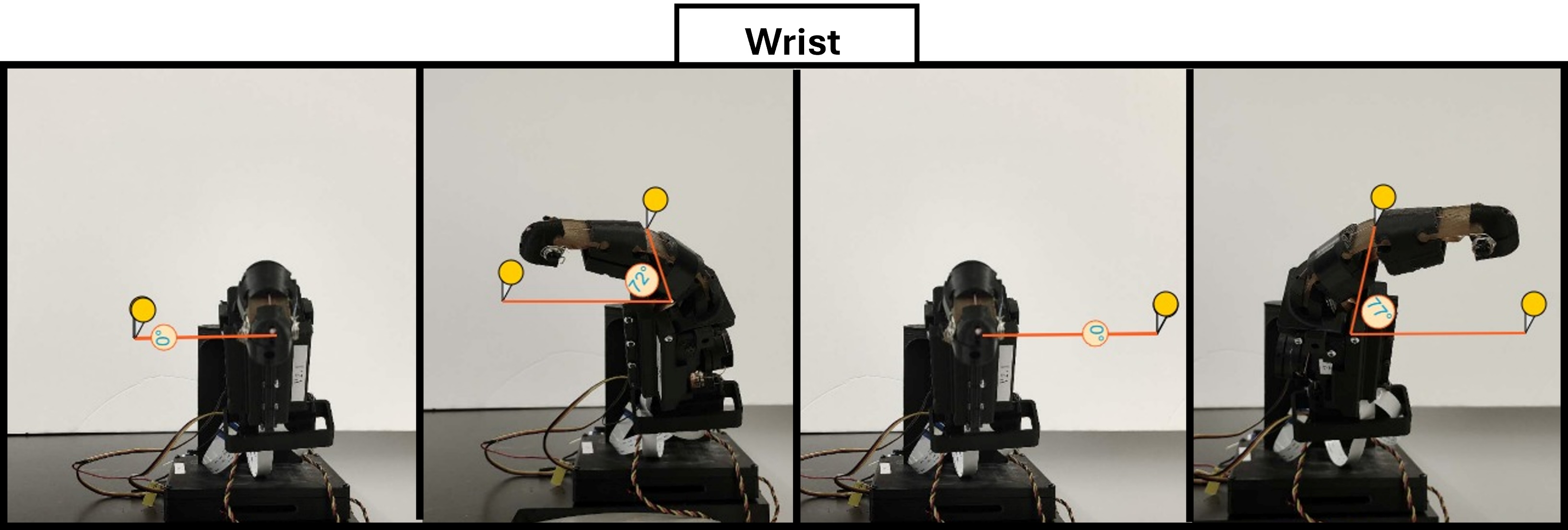}
    \caption{These tests drove the joints to their safe limits. Main and vertical axes are approximately 90deg, which is the full range of those servos. The wrist range of motion is approximately 150deg, which is limited from the full servo range of 180deg, due to cable binding.}
    \label{fig:range_of_motion}
\end{figure}

\section{MACHINE LEARNING EXPERIMENT} \label{sec:machine_learning_experiment}

\paragraph{Methodology}

\begin{figure}[h]
    \centering
    \includegraphics[width=.575\linewidth]{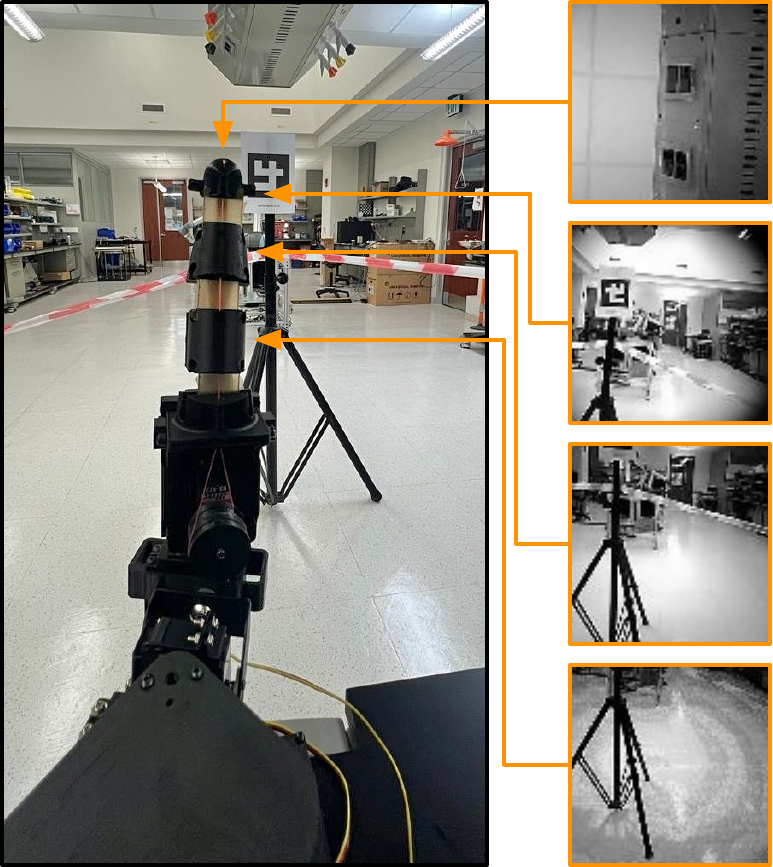}
    \caption{In our experiment, we use the four front-facing cameras and train the robotic libm to move from a random start position of the servos to a position where the target is in sight of one or more of the available cameras.}
    \label{fig:experiment_setup}
\end{figure}

As a proof of concept for the developed robotic limb in advanced machine learning tasks, we design a target-finding experiment.
In this task, the robotic limb adjusts its servo motors until a fixed target (Apriltag~\cite{olson2011apriltag}) is detected by any of the cameras positioned along its upper three joints (see Figure~\ref{fig:experiment_setup}).
This experiment employs two widely-adopted reinforcement learning algorithms: Proximal Policy Optimization (PPO)\cite{schulman2017proximal} and the Actor-Critic (AC) algorithm.
While PPO is optimal for continuous action spaces, AC is specifically tailored for discrete action spaces.
Both algorithms are trained in real-time using online data collection with minimal supervision.
Our PPO implementation is derived from a modified version of CleanRL\cite{huang2022cleanrl}, and the AC algorithm integrates Gaussian parameterization~\cite{witten1977adaptive,barto1983neuronlike,sutton2018reinforcement}.

To demonstrate how redundant cameras improve the robotic limb's robustness, we randomly kill some cameras by masking their image feed, using only the remaining cameras for detection.
The experiment comprises three settings: all cameras on, $1$ camera killed, and $1$-$2$ cameras killed.

For PPO, all four servo motors are adjusted simultaneously at each time step, whereas in AC, a single servo motor is adjusted per step.
The input state for both algorithms consists of two components: the angular positions of the four servos and the on/off status of the cameras.
PPO outputs four continuous values representing the adjustments to the servo positions (clipped to $[-36\degree, 36\degree]$), while AC uses an $8$-dimensional discrete action space, with each action corresponding to a $22.5\degree$ shift in a servo motor, either positive or negative.
The reward function assigns a value of $1$ when the target is spotted and $0$ otherwise.
To encourage exploration, the initial position of the robotic limb is randomized at the start of each episode.
In PPO, the episode ends when the target is detected or after a maximum of $5$ steps, preventing overfitting to environmental cues.
The policy is updated every $15$ episode, with a total of $3600$ steps runs per experiment.
In contrast, AC requires complete episode tracks for stable training, so episodes only end when the target is detected.
Each AC experiment consists of $250$ episodes.

We run both algorithms under each camera-kill setting for $10$ repetitions.
As a baseline, we conducted two $2$-hour Brownian motion tests: multiple servos (BMMS) moving simultaneously and a single servo moving at a time (BMSS), serving as baselines for PPO and AC, respectively.
The experiments use an NVIDIA Jetson Orin Nano Developer Kit\footnote{See \url{https://www.nvidia.com/en-us/autonomous-machines/embedded-systems/jetson-orin/}}, with the neural network trained in PyTorch~\cite{paszke2019pytorch} using CUDA 12.2 on a 1024-core NVIDIA Ampere GPU.

\begin{table*}[tb]
\centering
\vspace{.5em}
\setlength{\tabcolsep}{4mm}{}
\renewcommand\arraystretch{1.15}
\caption{Comparison of Algorithm Performances Across Camera-Kill Settings}
\label{tab:result_table}
\small
\begin{tabular}{|c|cccc|}
\hline
\multirow{2}{*}{Camera-Kill Setting} &
  \multicolumn{4}{c|}{Average Episode Lengths and Standard Deviation (SD)} \\ \cline{2-5}
 &
  \multicolumn{1}{c|}{BMSS} &
  \multicolumn{1}{c|}{BMMS} &
  \multicolumn{1}{c|}{PPO} &
  AC \\ \hline
Kill 0 &
  \multicolumn{1}{c|}{44.64 $\pm$ 47.44} &
  \multicolumn{1}{c|}{16.43 $\pm$ 15.35} &
  \multicolumn{1}{c|}{1.06 $\pm$ 0.17} &
  15.65 $\pm$ 3.37 \\
Kill 1 &
  \multicolumn{1}{c|}{59.34 $\pm$ 71.98} &
  \multicolumn{1}{c|}{20.89 $\pm$ 21.54} &
  \multicolumn{1}{c|}{2.32 $\pm$ 0.57} &
  18.06 $\pm$ 6.93 \\
Kill 1-2 &
  \multicolumn{1}{c|}{65.13 $\pm$ 76.57} &
  \multicolumn{1}{c|}{26.47 $\pm$ 28.41} &
  \multicolumn{1}{c|}{2.83 $\pm$ 0.74} &
  20.23 $\pm$ 17.47 \\ \hline
\end{tabular}
\end{table*}

\begin{figure}[htb]
    \centering
    \includegraphics[width=\linewidth]{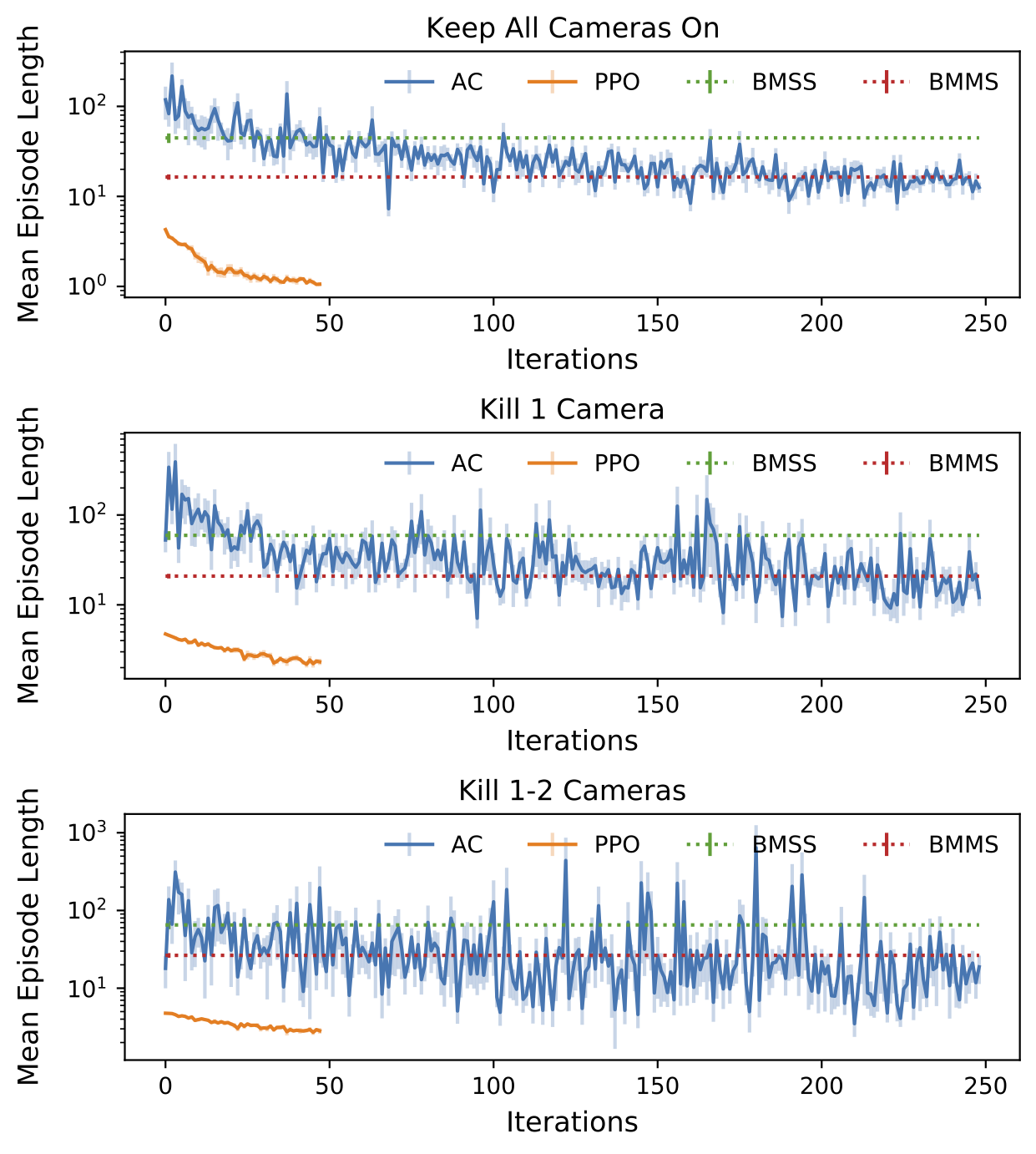}
    \caption{
        Performance of PPO over iterations and AC over episodes under varying camera-kill settings with standard error bars.
        Both show dramatic performance improvements in a relatively short time.
        By the end, both algorithms significantly outperform ($p < 0.01$) their respective baselines: Brownian noise with multiple (all) servos (BMMS) and a single servo (BMSS).
    }
    \label{fig:episode_lengths}
\end{figure}

\paragraph{Results}
The experimental results under the three camera-kill settings are shown in Figure~\ref{fig:episode_lengths}.
As expected, PPO converges faster and with less variability between runs than AC, primarily due to its capped episode length of five steps and the ability to adjust multiple joints simultaneously, which enables the robotic limb to locate the target more efficiently.
For PPO, the number of steps is bounded between $1$ and $5$, corresponding to the optimal policy and the truncated episode length, respectively.
In contrast, the number of steps is unbounded for AC, and very long episodes were observed at the beginning of training, skewing the averages shown in the learning curve.
This result occurred more frequently when the fingertip camera was killed, as larger joint movements were needed to locate the target.
Across all camera-kill settings, later phases of training were dominated by short episodes for both PPO and AC, with both significantly outperforming Brownian noise baselines in their final iterations ($p < 0.01$ using Student's t-tests with Holm-Bonferroni corrections).

Table~\ref{tab:result_table} shows the performance of both algorithms under different camera-kill settings, including the average episode length to find the target and the number of iterations/episodes needed for convergence.
As expected, performance worsened slightly as more cameras were disabled.
For PPO, the average steps to detect the target increased by $1.26$ and $1.77$ as the number of disabled cameras increased from $0$ to $1$ and $1$-$2$.
For AC, the impact of camera-kill was more pronounced, with increases in episode length of $2.41$ and $4.58$, but still achieves convergence.
Both PPO and AC demonstrated robust learning, even in the most challenging scenario where only half cameras are functional.
The entire training process lasted $8$ days, during which the robotic limb experienced only one mechanical failure, which occurred in a part designed for easy replacement.
Overall, the robotic limb displayed strong mechanical robustness throughout the experiments.

\paragraph{Discussion}
These experiments show that the algorithms are able to converge to optimal behaviour under the assumption violations introduced by the robotic platform.
The lack of any divergent runs reinforces this result.
Additionally, the algorithms effectively leverage redundant sensors to compensate for loss, completing the task even when most sensors are masked.
This resilience highlights the system's robustness in adapting to sensory degradation and confirms its capability to train state-of-the-art machine learning models under challenging conditions.
These results fulfill the primary objective outlined in Section~\ref{sec:introduction}.

\section{CONCLUSIONS AND FUTURE WORK} \label{conclusions_and_future_work}

\paragraph{Conclusions}
This paper details our design, build, and test of a robust hybrid hard/soft robotic limb for long period machine learning experiments.
The limb in question has $6$ cameras, $4$ servo motors, and is a combination of a soft tentacle-like outer limb with 2-directional shoulder and wrist hard inner limb.
We show that this limb has full range of motion without damage, using the hard inner limb for consistent degrees of freedom and soft outer limb for damage-resistant sensing and interaction with the world.
We show that this limb is robust to AI experiments by conducting a target-finding task and demonstrating convergence.

This limb demonstrates a complete capability for moving curiosity-driven baby robots out of controlled, limited, and supervised lab environments into environments in which the algorithms can start to flourish.
We already demonstrated the reduced need for supervision, even with full range of motion.
We believe AI/ML experiments on such a robotic limb enables the next steps toward artificial generalized intelligence on robotics that is not limited by the design assumptions built into those robots by human designers.

\paragraph{Future Work}
Wire management and binding is a complex issue. As can be seen in the images of the arm, the wires are kept loose enough to avoid much binding, but that results in wires that can catch on obstacles in the environment. Our lessons learnt include:
\begin{itemize}
    \item The camera ribbon cable extension connectors are failure points, because these sliding result in the cables getting either bent or pulled from the connector
    \item The tendon routing and control needs to be more tightly managed to avoid eventual loss of control
    \item The partially managed wiring near the base of the arm is a failure point in more complex environments.
\end{itemize}
Future work entails improvements based on these lessons learnt to enable unsupervised learning in complex environments.
Subsequent to resolving these issues, future experiments are planned to involve contact and manipulation for the robotic limb in more complex environments.

\section*{ACKNOWLEDGMENT}

This work was supported by the European Research Council (ERC, Advanced Grant Number 742870) and the Swiss National Science Foundation (SNF, Grant Number 200021 192356).

\bibliographystyle{IEEEtran}
\bibliography{dylan,main}

\end{document}